\documentclass[journal]{IEEEtran}

\usepackage{cite}
\usepackage{comment}

\ifCLASSINFOpdf
\usepackage[pdftex]{graphicx}
\else
\usepackage[dvips]{graphicx}
\fi

\usepackage{subfig}
\usepackage{amsmath}

\usepackage{url}

\usepackage[colorinlistoftodos]{todonotes}

\usepackage{array}
\usepackage{ragged2e}
\usepackage{adjustbox}
\usepackage[ruled,linesnumbered,vlined]{algorithm2e}

\begin{document}

\title{Conceptual Game Expansion}
\author{\IEEEauthorblockN{Matthew Guzdial\IEEEauthorrefmark{1},
Mark O. Riedl\IEEEauthorrefmark{2},
}
\IEEEauthorblockA{\IEEEauthorrefmark{1}Department of Computing Science, Alberta Machine Intelligence Institute, University of Alberta, Edmonton, AB T6G 0T3, Canada}
\IEEEauthorblockA{\IEEEauthorrefmark{2}School of Interactive Computing,
Georgia Institute of Technology, Atlanta, GA 30332, USA}
\IEEEauthorblockA{emails: guzdial@ualberta.ca, riedl@cc.gatech.edu}
}

\markboth{IEEE Transactions on Games}%
{Conceptual Game Expansion}

\maketitle

\begin{abstract}
Automated game design is the problem of automatically producing games through computational processes. 
Traditionally, these methods have relied on the authoring of search spaces by a designer, defining the space of all possible games for the system to author. 
In this paper, we instead learn representations of existing games from gameplay video and use these to approximate a search space of novel games. 
In a human subject study we demonstrate that these novel games are indistinguishable from human games in terms of challenge, and that one of the novel games was equivalent to one of the human games in terms of fun, frustration, and likeability. 
\end{abstract}


\IEEEpeerreviewmaketitle

\section{Introduction}

Video game design and development requires a large amount of expert knowledge. 
This skill requirement serves as a barrier that restricts those who might most benefit from the ability to make games, such as educators, activists, and those from marginalized backgrounds.
Researchers have suggested automated game design as a solution to this issue, in which computational systems produce games without major human intervention~\cite{treanor2012game}.
The promise of automated game design is that it could democratize game design by reducing this skill requirement barrier.
However, automated game design has relied upon encoding human design knowledge via authoring parameterized game design spaces, game components or entire games for a system to remix.
This authoring work requires expert knowledge and takes a large amount of time, which limits the applications of these automated game design systems~\cite{cook2016angelina}.

Machine learning (ML) has recently shown success at many tasks. 
One might consider applying machine learning to solve the automated game design knowledge authoring problem, learning the required knowledge from existing games instead of requiring human authoring.
However, modern machine learning requires large datasets with thousands of instances \cite{UnreasonableEffectivenessOfData}, and often underperforms or fails when confronted with contexts outside its original training sets.
Thus, it is insufficient to try to naively apply machine learning to solve this problem.

To address the shortcomings of traditional AI and ML approaches for automated game design, we explore the application of combinational creativity to this task.
Combinational creativity, also called conceptual combinational, is a particular type of cognitive process for recombining old knowledge to make something new, like combining the wings of a bird and the body of a horse for a flying horse or Pegasus \cite{Gagn1997,boden1998creativity}. 
There have been a number of computational attempts to recreate this process. 
We apply machine learning to learn representations of existing games and then use a novel, computational approximation of combinational creativity to recombine these representations to produce new games.

In this paper, we draw on novel machine learning methods to derive models of level design and game mechanics, and combine them in a representation we call a \textit{game graph}.
For training data, we draw on gameplay video as a representation of a player's experience with a game. 
We employ a combinational creativity approach called conceptual expansion to combine learned game graphs for three existing games.
These combinations represent novel game graphs, which we then then transform into playable games.
We evaluate these games in a human subject study comparing them to human-designed games and baseline games produced by other combinational creativity approaches.

\section{Related Work}
\label{sec:related-work}

In this section, we summarize prior related work on combinational creativity, machine learning focused on games and procedural content generation.
Notably we focus on the most relevant prior work, and especially prior work that relates to our own in technique or subject matter.

\subsection{Combinational Creativity}

Combinational creativity algorithms attempt to approximate the human ability to recombine old knowledge to produce something new.
The input of these algorithms tends to be a pair of graphs with the output of a new graph with features that are a combination of features from the two input graphs.
There are two core parts of traditional combinational approaches: mapping and constructing the final combination.
Mapping determines what components from the old knowledge may be relevant to approximate valuable new knowledge and what components of the old knowledge might be combined (i.e. nodes and edges for graph input). 
Mapping typically requires some knowledge base to allow these algorithms to reason over common-sense knowledge.
How the final combination is constructed or chosen is the major difference among combinational creativity approaches.

Combinational creativity algorithms tend to have many possible valid outputs. 
Traditionally, this is viewed as undesirable, with general heuristics or hand-authored constraints designed to pick a single final combination \cite{fauconnier2001conceptual,ontanon2010amalgams}.
This limits the potential output of these approaches, we instead employ a domain-specific heuristic to find an optimal combination from a space of possible combinations.

We identify three prior combinational creativity approaches for deeper investigation, due to the fact that they are well-formed in domain-independent terms: conceptual blending, amalgamation, and compositional~adaptation.
It is worth noting that the final combination and the approach are referred to with the same words (e.g. conceptual blending produces conceptual blends).
We employ these three approaches as baselines during our human subject study.
In Figure \ref{fig:levelDesignComparisonFigure}, we give an example of a pair of input graphs representing two types of local Super Mario Bros. level design, a partially specified mapping, and example output of the three existing approaches and our approach: conceptual expansion.
In this simplified example, combinations of components are represented with both components in the same node. 
Conceptual expansion differs from the other approaches in this simple example by allowing for output with arbitrary sizes and combinations.

\subsubsection{Conceptual Blending}

Fauconnier and Turner \cite{fauconnier1998conceptual} formalized the ``four space" theory of conceptual blending. 
In this theory they describe four spaces that make up a blend: two \textit{input spaces} represent the unblended elements, input space points (or features) are projected into a common \textit{generic space} to identify equivalence, and these equivalent points are projected into a \textit{blend space}. In the blend space, novel structure and patterns arise from the projection of equivalent points.
Functionally, conceptual blending implementations attempt to retain as much information from the input as possible. 
In Figure \ref{fig:levelDesignComparisonFigure}, the conceptual blend output graph (``blend'') is the largest for this reason.
It combines all the mapped components and retains the unmapped components. 
This combination of mapped components is domain-dependent, for example Goguen and Harrell \cite{goguen2004style} give an example combining the words ``house'' and ``boat'' to create ``houseboat''. 
  
\subsubsection{Amalgamation}

Onta{\~n}{\'o}n designed amalgams as a formal unification function between multiple cases \cite{ontanon2010amalgams}. 
Similar to conceptual blending, amalgamation requires a knowledge base that specifies when two components of a case share a general form, for example \textit{French} and \textit{German} can both share the more general form \textit{nationality}.
Unlike conceptual blending, this shared generalization does not lead to a combination of components, but requires that only one of the two be present in an output amalgam. 
Essentially, amalgamation chooses between mapped components, with only one component from each present in the output.
For example, a \textit{red French car} and an \textit{old German car} could lead to an \textit{old red French car} or an \textit{old red German car}. 
Thus, the amalgam in Figure \ref{fig:levelDesignComparisonFigure} is much smaller than the blend, without combined components. 

\subsubsection{Compositional Adaptation}

\begin{figure}[tb]
	\centering
	\includegraphics[width=\columnwidth]{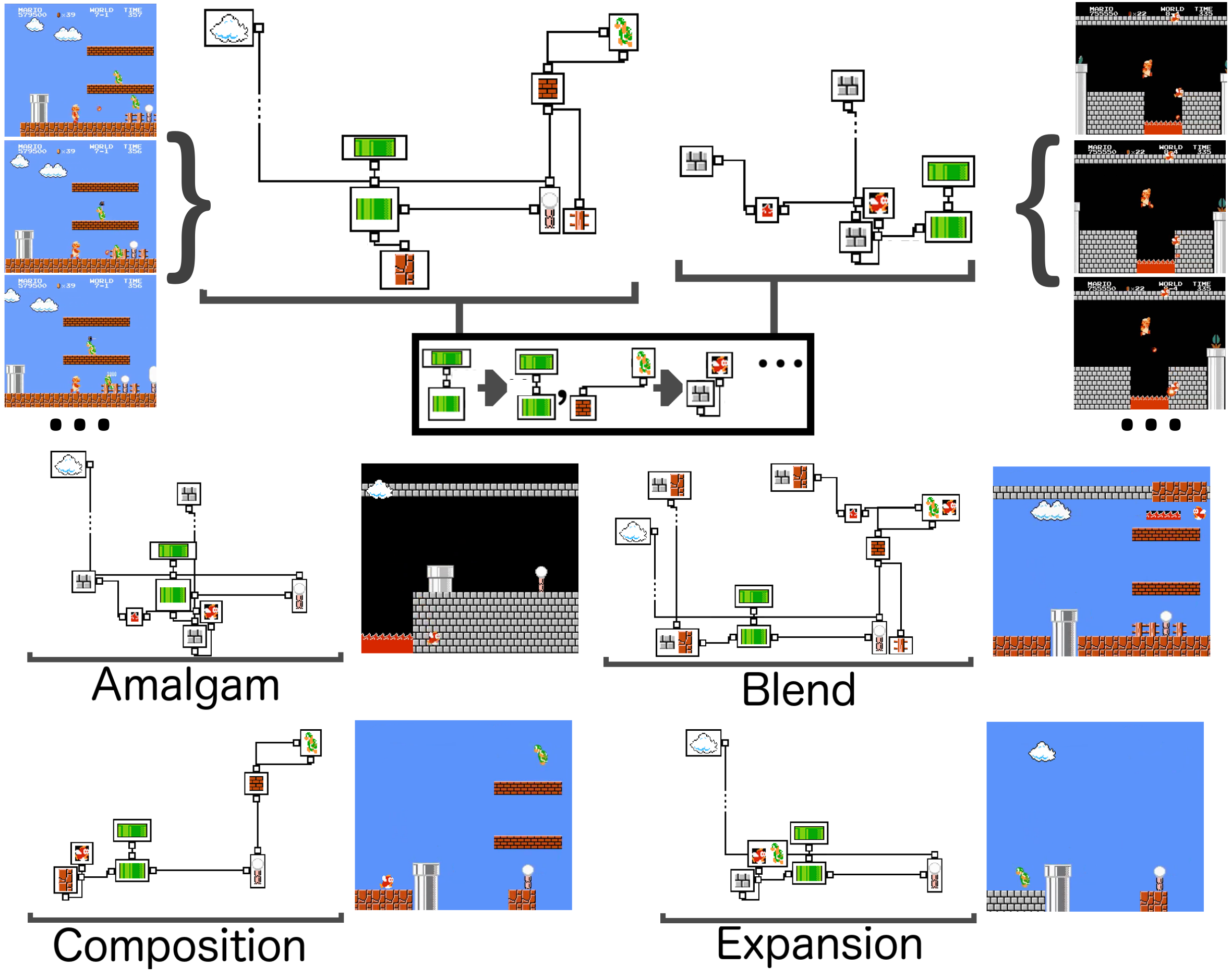}
	\caption{A comparison of the different combinational creativity approaches on graphs representing two different Super Mario Bros. level types, adapted from \cite{guzdial2018combinatorial}. The two input graphs at the top represent different types of local level designs for above ground (left) and boss/castle (right) levels, both are learned from frames as in \cite{guzdial2016toward}. These graphs and the authored mapping (the box), are fed to each of the combinational creativity approaches. Below we give a graph output of each approach and an example of a piece of level that graph could represent. }
	\label{fig:levelDesignComparisonFigure}
\end{figure}

Compositional adaptation arose as a CBR adaptation approach \cite{holland1989induction,fox2009exploring}, but has found applications in adaptive software \cite{mckinley2004taxonomy,eisenbach2007component}.
The intuition behind compositional adaptation is that input nodes can be broken apart and then connected to create novel structure based on their shared edges. 
In adaptive software this is sets of functions with given inputs and outputs that can be strung together to achieve various effects, which makes compositional adaptation similar to planning or a grammar when it includes a goal. 
However, it can also be applied in a goal-less way to generate sequences or graphs of components. 
Unlike amalgamation and conceptual blending, compositional adaptation does not require an explicit knowledge base by default. 
However, it is common to make use of a knowledge base to generalize components and their relationships in order to expand the set of possible combinations. 
A mapping is one way to represent this knowledge base, mapped components are considered equivalent, meaning that any edge connected to one node can be connected to another node in an output (composition) if the two nodes are mapped.
In Figure \ref{fig:levelDesignComparisonFigure}, the composition is smaller than the amalgam or blend without combined components. 
However, it could have been larger than either the amalgam or blend by incorporating more of the edges and nodes from the inputs.

\subsection{Procedural Content Generation}

\textit{Procedural content generation} is the umbrella term for systems that take in some design knowledge and output new content (game assets) from this knowledge. 
Approaches include evolutionary search, rule-based systems and instantiating content from probability tables \cite{hendrikx:2013:PCG,togeliusSurvey}. 

\subsection{PCGML}

Procedural content generation via machine learning (PCGML) \cite{summerville2017procedural} covers a range of techniques for generating content from models trained on existing content. 
Super Mario Bros. level generation has been the most popular domain thus far, with researchers applying techniques such as Markov chains \cite{snodgrass2016learning}, Monte-Carlo tree search \cite{summerville2015mcmcts}, long short-term recurrent neural networks \cite{SMBRNN}, autoencoders \cite{jain2016autoencoders}, generative adversarial neural networks \cite{volz2018evolving}, and genetic algorithms through learned evaluation functions~\cite{dahlskog}. 

Combinational creativity has rarely been applied to PCGML. 
However there has been increasing recent interest in the topic, particularly work by Sarkar and Cooper \cite{sarkar2020sequential,sarkar2020towards}, in which models are trained on multiple game levels simultaneously to approximate neural conceptual blending. 

The major differences between our research and the other work in the field of PCGML are (1) the source of training data and (2) the focus on producing entirely new games.
First, by focusing on gameplay video as the source of training data we avoid the dependence on existent corpora of game content, opening up a wider set of potential training sources. 
Second, our work focuses on the creative generation of novel games, while most prior PCGML techniques attempt to recreate content in the same style as the given training data.

\subsection{Automated Game Design}

In this section, we cover examples of automated game design: PCG approaches used to produce entire games.
Treanor et al. \cite{treanor2012game} introduced Game-o-matic, a system for automatically designing games to match certain arguments or micro-rhetorics \cite{treanor2012micro}. 
This process created complete, if small, video games based on a procedure for mapping these arguments to pre-authored mechanics. 
Cook et al. developed the ANGELINA system for automated game design and development \cite{cook2017angelina}. 
There have been a variety of ANGELINA system versions, with most focusing on a particular game genre.
Each version employed grammars and genetic algorithms to create game structure or rules \cite{cook2013mechanic}.

Existing work has implicitly or explicitly invoked computational creativity for automated game design.
Nelson and Mateas \cite{nelson2008recombinable} proposed a system to swap the rules of games to create new experiences. 
Gow and Corneli \cite{gow2015towards} proposed a similar system explicitly drawing on conceptual blending.
Nielsen et al. \cite{nielsen2015general} introduced an approach to mutate existing games expressed in the video game description language (VGDL) \cite{schaul2013video}. 
Nelson et al. \cite{nelson2016mixed} defined a novel parameterized space of games and randomly altered subsets of parameters to explore this design space.
More recent work has applied variational autoencoders to produce noisy replications of existing games, called World Models or Neural Renderings \cite{ha2018world,eslami2018neural}. 
This work does not attempt to create new games, but to recreate subsections of existing games for automated game playing agent training purposes. 

Osborn et al. \cite{osborn2017automated} proposed automated game design learning, an approach employing emulated games to learn a representation of the game's structure \cite{osborn2017automatic} and rules \cite{summerville2017charda}. 
This approach is most similar to our work in terms of learning a model of a game's structure and rules. 
However, this approach depends upon access to a game emulator and has no existing process for creating novel games.

\begin{figure}[tb]
	\centering
	\includegraphics[width=\columnwidth]{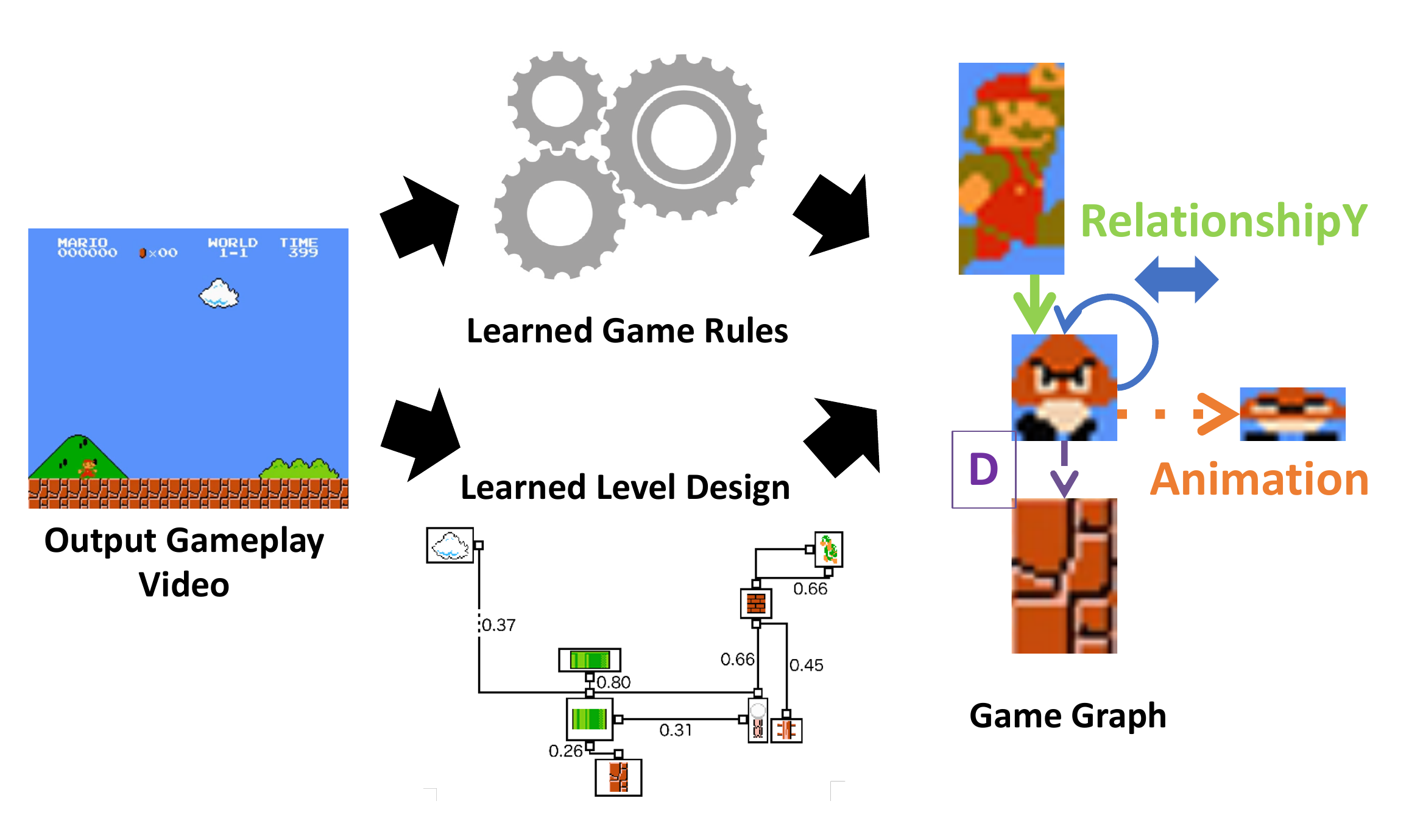}
	\caption{A visualization of the entire process for learning game graphs.}
	\label{fig:fullGameGraphLearning}
\end{figure}

\section{Game Graph Representation}

To generate new video games we employ a novel representation we call a game graph, which encodes both level-design information and game ruleset information.
We visualize the process of learning a game graph in Figure \ref{fig:fullGameGraphLearning}.
We learn a representation of the game's level design and rules from gameplay video and then combine this information into a graph.
In this section, we discuss how we learn the two constituent parts and construct a game graph for a given game.
Unfortunately, each learned game graph in this work had dozens of nodes and over a thousand edges, making a complete visualization impossible.
To create new games, we combine components from multiple game graphs, which we discuss in the next section.

The process for learning the input game graphs is as follows: we take as input gameplay video and a spritesheet. 
A spritesheet is a collection of all of the images or sprites in a game, including all background art, animation frames, and components of level structure. 
In this paper, we used two videos of approximately two minutes each (focused on a single level) and a spritesheet put together by fans for each game.
We run image processing with OpenCV \cite{Pulli:2012:RCV:2184319.2184337} on the video with the spritesheet to determine where, and what sprites occur in each frame. 
We refer to this as a \emph{level chunk}, an approximation of the frame with game entities instead of raw pixels. 
Then, we learn a model of level design and a ruleset for the game from this input. 
We then use the models of level design and game ruleset to construct a game graph. 
In the following subsections we review the pertinent aspects of the techniques we use to learn models of level design and game rulesets, and how we create game graphs from the output of these techniques. 
We ran this process on three games to derive game graphs: Super Mario Bros., Mega Man, and Kirby's Adventure.


\subsection{Level Design Learning}

In this section, we summarize our process for learning level design knowledge, which is covered in greater detail in \cite{guzdial2016toward,guzdial2016game}. 
At a high level this technique derives a hierarchical graphical model or Bayesian network that represents probabilities over local level structure.
The model is made up of three types of observed nodes derived directly from the level chunks. 
$G$ nodes represent geometric shapes of a sprite type $t$, like the long flat sections of ground in the frames in Figure \ref{fig:marioRuleExample}. 
$D$ nodes represent relative positions between $G$ nodes. 
$N$ nodes represent counts of sprites in a particular level chunk.

Our level design model has two hidden nodes, which are approximated via $K$-means clustering with $K$ estimated via the distortion ratio \cite{Pham01012005}. $S$ nodes, which are approximated by clustering $G$ and $D$ nodes, represent styles of shapes of sprite type $t$, in terms of where they occur relative to other shapes ($D$) and their geometry ($G$). 
$L$ nodes, which are approximated by clustering $S$ and $N$ nodes, represent styles of level chunks, in terms of the types of sprites that occur within them ($N$) and the styles of those sprites ($S$). 

The goal of this process is to approximate the probability distribution $P(g_{s_i}|g_{s_j}, d)$ for each $L$ node. 
This represents the probability of a certain shape of sprites ($g_{s_i}$) of style ($s_i$) at a relative position $d$ to another shape of sprites ($g_{s_j}$) of a particular style ($s_j$). 
For example, the probability of seeing a goomba sprite, a common Mario enemy, just above a rectangle of ground sprites. 
Starting from a level chunk with one sample $g$ value, we can query this probability distribution until it satisfied a sampled $N$ node value to generate new level chunks of a particular level chunk type.

We learn a probabilistic ordering of $L$ nodes to generate full levels by processing the sequence of frames as a sequence of $L$ nodes. 
We find the level chunk value associated with each frame, and find the $L$ node that represents that type of level chunk.
This gives us a sequence of $L$ nodes for each level expressed in our video footage, which we merge with the fuzzy graph merge algorithm \cite{foley1999fuzzy}.
We call this final directed graph a \emph{level graph}, which is a graph of $L$ nodes with weights on its edges represting the probability of transitioning between different types of level chunks.
We walk these edges to create a sequence of $L$ nodes, which we can covert into level chunks and string together into a complete level.

\subsection{Ruleset Learning}

\begin{figure}[tb]
	\centering
	\includegraphics[width=\columnwidth]{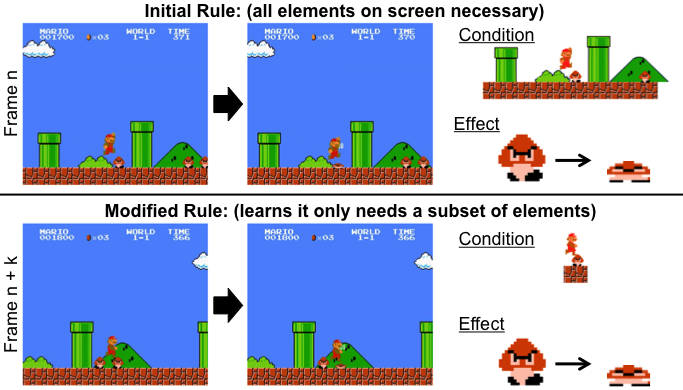}
	\caption{A visualization of two pairs of frames and an associated engine modification. Reproduced from \cite{guzdial2018automated}.}
	\label{fig:marioRuleExample}
\end{figure}

In this section, we summarize the approach to learn rules from gameplay video as originally described in \cite{guzdial2017game}. 
This technique re-represents each gameplay video frame as a list of conditional facts that are true in that frame. 
The fact types are as follows:

\begin{itemize}
\item \textbf{$Animation$: } Each animation fact tracks a particular sprite seen in a frame by its name, width, and height. 
\item \textbf{$Spatial$: } Spatial facts track spatial information, the $x$ and $y$ locations of sprites on the screen.
\item \textbf{$RelationshipX/RelationshipY$: } The RelationshipX and RelationshipY facts track the relative positions of sprites to one another in their dimension.
\item \textbf{$VelocityX/VelocityY$: }The VelocityX and VelocityY facts track the velocity of entities in their dimension.
\item \textbf{$CameraX$/$CamereaY$: }Tracks the camera's position in their dimension.
\end{itemize}
\noindent
The algorithm iterates through pairs of frames, using its current (initially empty) ruleset to attempt to predict the next frame. 
When a prediction fails it begins a process of iteratively updating the ruleset by adding, removing, and modifying rules to minimize the distance between the predicted and actual next frame. 
We visualize adding and later modifying a rule in Figure \ref{fig:marioRuleExample}.
The rules are constructed with conditions and effects, where conditions are a set of facts that must exist in one frame for the rule to fire, and effects are a pair of facts where the second fact replaces the first when the rule fires. 

\subsection{Game Graph Construction}

The output of the level design model learning process is a probabilistic graphical model and probabilistic sequence of nodes. 
The output of the ruleset learning process is a sequence of formal-logic rules. 
We take knowledge from both of these to construct a game graph. 
This represents a learned, graphical representation of a game.

The construction of a game graph is straightforward. 
Each sprite in a spritesheet for a particular existing game becomes a node in an initially unconnected graph. 
We then add all the knowledge from the level design model and ruleset representations as edges to these nodes.
There are nine distinct types of edges that contain different types of knowledge:

\begin{itemize}
\item \textbf{G: } Stores the value of a $G$ node from the level design model, represented as the $x$ and $y$ positions of the shape of sprites, the shape of sprites (represented as a matrix), and a unique identifier for the $S$ and $L$ node that this $G$ node value depends on. 
This edge is cyclic, pointing to the same component it originated from. 
We note one might instead store this information as a value on the node itself, but treating it as an edge allows us to compare this and other cyclic edges to edges pointing between nodes.
\item \textbf{D: } Stores the value of a $D$ node connection, as a vector with the relative position, the probability, and a unique identifier for the $S$ and $L$ node that this $D$ node value depends on. 
This edge points to the equivalent node for the sprite this $D$ node connection connected to in the training data.
\item \textbf{N: } Stores the value of an $N$ node, which is a value representing a particular number of this component that can co-exist in the same level chunk and a unique identifier for its $L$ node. 
This edge is cyclic.
\item \textbf{Rule condition: } Stores the value of a particular fact in a particular rule condition, which includes the information discussed for the relevant fact type and a unique identifier for the rule it is part of.
This edge can be cyclic or can point to another component (as with Relationship facts).
\item \textbf{Rule effect: } Stores the value of a particular rule effect, which includes the information for both the pre and post facts and a unique identifier for its rule. 
This edge can be cyclic or can point to another component.
\end{itemize}

\begin{figure}[tb]
	\centering
	\includegraphics[width=\columnwidth]{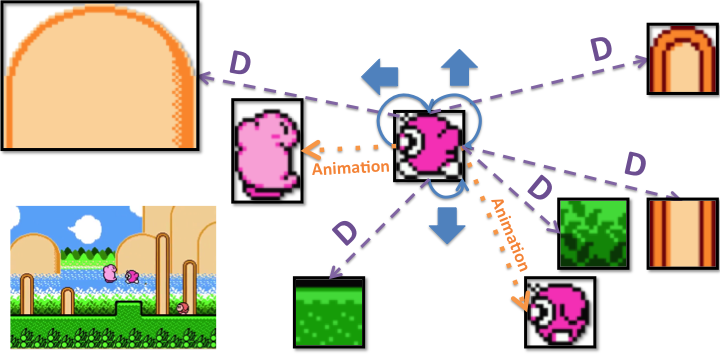}
	\caption{A subset of the game graph centered on one Waddle Doo enemy sprite from Kirby's Adventure, along with a relevant gameplay frame. Reproduced from \cite{guzdial2018automated}.}
	\label{fig:componentGraph}
\end{figure}

A prior game graph representation did not include the level graph data, the probabilistic sequencing data that allowed for entire levels to be constructed instead of level chunks \cite{guzdial2018automated}.
We add a node for each $L$ node and the following four edges in order to capture this information.

\begin{itemize}
\item \textbf{Level Chunk Type: } This simply stored the id of this level chunk category, which is an arbitrary but unique id generated during the level chunk categorization process.
Notably this id is related to the information stored in the \textbf{D}, \textbf{G}, and \textbf{N} edges to identify their associated $L$ node. 
Thus generated $L$ nodes can be associated with this level chunk sequence information when generating levels from game graphs.
This edge is cyclic, pointing to the same component it originated from. 
\item \textbf{Level Chunk Repeats: } Stores a minimum and maximum number of times this level chunk type can repeat in a sequence.
This edge type is cyclic.
\item \textbf{Level Chunk Position: } Stores a float value specifying the average, normalized position where this level chunk tends to be found in a sequence of level chunks.
This information is not used during level generation, but is a helpful feature for mapping similar level chunk category nodes.
This edge is cyclic.
\item \textbf{Level Chunk Transition: } Stores a pointer to another level chunk category node and the probability of taking that transition.
However, notably this edge stores insufficient data to recreate the entire level graph. 
Instead this solves the level chunk sequence generation problem with a Markov chain-like representation, sampling from possible transitions probabilistically until it hits a node without any outgoing transitions.
This edge points from this node to the associated level chunk category node.
\end{itemize}

We visualize a small subsection of a final game graph focused on the Waddle Doo enemy from Kirby's Adventure in Figure \ref{fig:componentGraph}.
The cyclic arrows in blue represent rule effects that impact velocity (we do not include the full rule for visibility). 
The orange dotted arrows represent rule effects that impact animation state. 
The dashed purple arrows represent $D$ node connection edges. 
The actual game graph is much larger, with dozens of nodes and thousands of edges. 
It contains all information from the learned level design model and game rulesets. 
That is, a playable game can be reconstructed from this graph. 
This is a large, dense representation, with each of the three game graphs we constructed having thousands of edges. 
The graph structure can be manipulated in order to create new games by adding, deleting, or altering the values on edges.
Small changes allow for small variations (e.g. doubling Mario's jump height by tweaking the values of two edges, one for the starting y velocity of the jump and another to begin the jump's deceleration).
Larger changes could then represent entirely distinct games.

\section{Conceptual Expansion}

The size, complexity, and lack of uniformity of the game graph representation makes them ill-suited to generation approaches that rely on statistical machine learning.
Instead we apply conceptual expansion to these game graphs. 
Conceptual expansion is a parameterized function that defines a space of possible output combinations.
We define the conceptual expansion function as:
\begin{equation} \label{eq1}
  \mathrm{CE^X}(F,A) = a_1*f_1+a_2*f_2+ ... +a_n*f_n
\end{equation}

\noindent
Where $F=\{f_1...f_n\}$ is the set of all mapped features and $A=\{a_1, ... a_n\}$ is a filter representing what of and what amount of mapped feature $f_i$ should be represented in the final conceptual expansion. 
$X$ is the concept that we are attempting to represent with the conceptual expansion.
Thus, applying equation~\ref{eq1}, $CE^X$ is a final node in the new game graph, $f_1$, $f_2$... $f_n$ are existing nodes and their associated edges from our knowledge base, and $a_1$, $a_2$... $a_n$ are the filters on these nodes, which determine what edges are present in the final output node and how they are transformed.
A particular $a_i$ determines what edges from $f_i$ are added to the final expanded node $CE^{X}$ and how those edges are altered from the original edge in $f_i$. 
In this particular implementation, we fix the number of final nodes, but this is not a requirement of the approach. 
Our new game graph is a set of conceptual expansions ($CE^{X}$) of each node ($X$) of that new graph. Essentially, each node of our new graph represents a combination of an arbitrary number of nodes from the existing graphs.

As an example, imagine that we do not have Goombas (the basic Mario enemy seen in Figure \ref{fig:marioRuleExample}) as part of an existing game graph. 
Imagine we are trying to recreate a Goomba node with conceptual expansion ($CE^{\mathtt{Goomba}}$), and we have the Waddle Doo node as seen in Figure \ref{fig:componentGraph} mapped to this Goomba node.
In this case there may be some edges from this node we want to take for the Goomba node such as the velocity rule effect to move left and fall, but not the ability to jump. 
In this case one can encode this with an $a_{waddle doo}$ that filters out jump (e.g. $a_{waddle doo}$ = [1,0...]). 
One can alter $a_{waddle doo}$ to further specify one wants the Goomba to move half as fast as the Waddle Doo. 
One might imagine other $f$ and corresponding $a$ values to incorporate other information from other game graph nodes for a final $CE^{Goomba}(F,A)$ that reasonably approximates a true Goomba node.
We demonstrated that conceptual expansion could be employed to more successfully recreate existing games than other standard automated game generation approaches in \cite{guzdial2018automated}.
However, we had not yet determined how human players would react to games produced by conceptual expansion, which we explore in this paper.

At a high level, conceptual expansion creates a parameterized search space from an arbitrary number of input game graphs. 
One can think of each aspect of a game level design or ruleset design as a dimension in a high-dimensional space. 
Changing the value of a single dimension results in a new game different from the original in some small way. 
Input game graphs provide a learned schematic for what dimensions exist and how they relate to each other.  
With two or more game graphs, one can describe new games as interpolations between existing games, extrapolations along different dimensions, or alterations in complexity. One can then search this space to optimize for a particular goal or heuristic, creating new games. 
One can then process these new game graphs back into level design models and rulesets to create new, playable games.

Conceptual expansion over game graphs has two steps. 
First, we derive a mapping, which gives us the initial $a$ and $f$ values for each expanded node. 
This is accomplished by determining the best $n$ nodes in the knowledge base according to some distance function. 
For the second step we make use of a greedy hill-climbing approach we call \textit{heuristic-driven conceptual expansion search}.

\label{ref:ce}
\IncMargin{1em}
\begin{algorithm}[tb]
\footnotesize
\SetKwData{MaxExpansion}{maxE}\SetKwData{Graph}{initGraph}\SetKwData{Mapping}{m}\SetKwData{Score}{score}\SetKwData{TotalData}{data}\SetKwData{MaxScore}{maxScore}\SetKwData{Improving}{improving}\SetKwData{Visited}{v}\SetKwData{Neighbor}{n}\SetKwData{NeighborScore}{s}\SetKwData{OldMax}{oldMax}

\SetKwFunction{GetDefaultExpansion}{ExpansionFromInit}\SetKwFunction{GetNeighbor}{GetNeighbor}\SetKwFunction{Heuristic}{Heuristic}\SetKwFunction{Max}{max}
\SetKwInOut{Input}{input}\SetKwInOut{Output}{output}
\Input{a partially-specified game graph $initGraph$, and a mapping $m$}
\Output{The maximum expansion found according to the heuristic}
\BlankLine

\MaxExpansion$\leftarrow$ \GetDefaultExpansion{\Graph}+\Mapping\;
\MaxScore$\leftarrow$ 0\;
\Improving$\leftarrow$ 0\;

\While{\Improving $<$ 10}{
  \Neighbor$\leftarrow$ \MaxExpansion.\GetNeighbor{}\;
    \BlankLine 
    \NeighborScore$\leftarrow$ \Heuristic{\Neighbor, \Graph}\;
    \OldMax$\leftarrow$ \MaxScore\;
    \MaxScore, \MaxExpansion $\leftarrow$ \Max{$[$\MaxScore, \MaxExpansion$]$, $[$\NeighborScore, \Neighbor$]$}\;
    \eIf{\OldMax $<$ \MaxScore}{\Improving$\leftarrow$\Improving+1\;}{\Improving$\leftarrow$0\;}

}

return \MaxExpansion\;
\caption{Heuristic-Driven Conceptual Expansion Search}\label{algo_goalBasedConceptualExpansionSearch}
\end{algorithm}\DecMargin{1em}
\normalsize

We include the pseudocode for our search algorithm in Algorithm \ref{algo_goalBasedConceptualExpansionSearch}.
The input to this process is a partially-specified game graph ($initGraph$) which defines the basic structure of the conceptual expansion (e.g. the number of nodes) and a mapping ($m$).
This mapping determines what components from our input graphs are initially ``mapped'' to each node in our initial game graph (e.g. the values of $F$ for that node).
On line one, the function ``ExpansionFromInit'' creates an initial set of nodes from the $initGraph$. 
For each expanded node it fills in all its $a$ and $f$ values according to a normalized mapping in which the best mapped component has $a$ values of 1.0, and the $a$ values of the remainder follow based upon their relative distance. 
We then begin a greedy search process with the operators from \cite{guzdial2018automated}, searching ten randomly sampled neighbors at each step, and stopping when a local maxima is reached.
We discuss our heuristic in the next section.

\section{Spritesheet-based Game Generation}

One issue at this point is how to construct new visuals for any generated games, as the given approach will output novel game graphs but where each entity is just a rectangle with a unique id \cite{guzdial2018automated}. 
It is common practice for artists to make spritesheets for games that do not exist. 
If we could alter the conceptual expansion generation of game graphs to target a particular spritesheet and make a game to match it, that would solve this issue.
Further, this limits the problem from one of creating any possible video game to creating games that match a particular spritesheet, making the evaluation much simpler.

\begin{figure}[tb]
	\centering
	\includegraphics[width=2in]{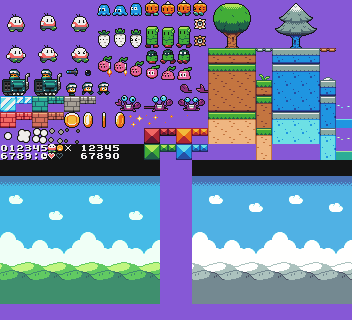}
	\caption{The GrafxKid ``Arcade Platformer Assets'' spritesheet used in this work.}
	\label{fig:grafxkidSpritesheet}
\end{figure}

In this section, we cover the pipeline for the spritesheet-based conceptual expansion game generator, the construction of what we call a proto-game graph from a spritesheet, the mapping from the three existing game graphs of Super Mario Bros., Mega Man, and Kirby's Adventure onto that spritesheet, and the heuristic we designed to represent value, surprise, and novelty.
We note that for the purposes of this paper we used the spritesheet in Figure \ref{fig:grafxkidSpritesheet}.

\subsection{Proto-Game Graph Construction}

Just by looking at a spritesheet a human designer can imagine possible games it could represent.
It stands to reason that constructing a game graph-like representation from a spritesheet could prove helpful as an analogue to this process.
Towards this end we developed a procedure to construct what we call a ``Proto-Game Graph'' from the spritesheet. 
It is common practice for visually similar sprites in a particular spritesheet to be related in terms of game mechanics and/or level design. 
Given this intuition we employ one round of single-linkage clustering on the individual sprites of a spritesheet in an attempt to automatically group visually similar sprites.
First, we represent each sprite as a bag of its 3x3 pixel features, given that bag of features strategies tend to perform well on image processing tasks even compared to state-of-the-art methods and we lack the training data for these methods \cite{brendel2019approximating}. 
We then construct a simple distance function as the count of the disjoint set of these features normalized by the sprite sizes.

We treat all clusters as a single game graph node. 
This gives an initial, underspecified game graph that we call a proto-game graph.
The proto-game graph technically has some mechanical (animation fact) and level design ($G$ node) information, but does not represent a playable game and could not be used to generate levels.
However, by having it in the same representation as a true game graph we can treat it as one for the purposes of constructing a mapping.

We identify the game graph nodes that should be used as the player of the game, represented as a simple boolean value on each node.
This is necessary as the player is identified for the three existing, learned game graphs. 
However, in order to minimize the potential for this choice to impact the creativity of the output all the baselines (including human-designed games) were also informed what sprites to treat as the player.

\subsection{Proto-Game Graph Mapping}

The first step of any combinational creativity approach is to determine a mapping.
In this case the nodes of the existing game graphs need to be mapped onto the nodes of the proto-game graph.
To accomplish this each node of all three of the existing game graph is mapped onto the closest node in the proto-game graph. 
We use an asymmetrical Chamfer distance metric as in \cite{guzdial2018automated}, which means that it does not matter that the proto-game graph nodes are under-specified. 
This distance function varies from [0,1] and we only allow mappings where the distance between the two nodes is $<$1, ensuring they have some similarity.
In the case that this mapping leaves some nodes in the proto-game graph without any existing game graph nodes mapped to them, which would leave them empty and mean that they had no chance to appear in any generated games, we map each of these empty proto-game graph nodes to its closest existing game graph node (flipping the direction of the distance function). 

We separately handle the $L$ nodes, nodes pertaining to the camera, and to the concept of nothing (``None''), used to represent empty game entities in the ruleset learning process, as they are not derivable from a spritesheet.
We add nodes for the Camera and None entities to the proto-game graph, due to the fact that all of the existing game graphs also have one of each of these specialized nodes. 
For the remaining unmapped existing game graph nodes we run K-medians clustering with $k$ estimated with the distortion ratio \cite{Pham01012005}. 
For the spritesheet proto-game graph used in this paper this led to five clusters of level chunk category nodes.
We map each cluster to a single node, meaning five final level chunk category nodes appeared for this spritesheet proto-game graph.

This mapping is used to derive an initial conceptual expansion for the conceptual expansion search process.
We also used it as the mapping for the three combinational creativity baselines (amalgamation, conceptual blending, and compositional adaptation) in the human subject study described below.

\subsection{Heuristic}

Our goal for the output games was that they be considered creative and unique.
Thus, for the heuristic we drew on the three-part model of creativity identified by Boden \cite{boden1998creativity}: novelty, surprise, and value.

For novelty we employ the minimum Chamfer distance to compare the current game graph to all existing game graphs in a knowledge base.
The knowledge base in this case includes the three original game graphs, and any previously generated game graphs created by the conceptual expansion game generator.
Intuitively, this can be understood as a measure of how dissimilar a particular game graph is compared to the most similar thing the system has seen before. 

Surprise is difficult to represent computationally given that it relies on some audience's expectations being contradicted \cite{boden1998creativity}.
We also wanted a metric that wouldn't just correlate exactly with novelty, but would represent a distinct measure.
Thus we first built a representation of audience expectations by constructing what we call \emph{Normalized Mapping Magnitude Vectors}.
One of these vectors is constructed for each of the three original game graphs, by mapping each node of each game graph to its closest node in the two remaining graphs. 
From this we collected the number of times each of the remaining two graph's nodes had been mapped, sorted these values and then normalized them.
This gives us three one-tailed distributions.
These essentially reflect how a person seeing one of these games would relate it to the two games they had already seen, or this hypothetical person's expectations.
To determine the surprise for a novel game graph the system constructs the same distribution by finding the mapping counts from the current graph to the existing graphs (including any graphs already produced by the system). 
We compare the distributions by cutting the tails so that they are equal, normalizing once again, and then directly measuring the distance between the two values at each position. 
Given these are normalized vectors, the max difference between the two would be 2, and thus we divide this number by 2.0 to determine the surprise value compared to each existing game.
We take the minimum of these as the surprise value.

Value is the final creativity component to represent in this heuristic. 
Value in games is difficult to evaluate objectively \cite{sweetser2005gameflow,bernhaupt2007methods}.
Given the focus of this paper is not on this unsolved problem we instead focused on only one aspect of a video game's value, which is more easily represented computationally: challenge.
However, the notion of automatically reflecting human experience across a level is another non-trivial problem and a current research area in terms of \textit{automated playtesting}~\cite{zook2014automatic,zook2015monte,holmgaard2018automated}.

We decided on an A* agent simulation approach.
Specifically, we looked at how an A* agent using the generated rules performed.
However, instead of only using the coarse measure of whether or not the agent could complete a level chunk, we collect three summarizing statistics from the A* agent's search across a given level chunk.
Specifically we collect the max distance the A* agent traversed, normalized to the width of the level chunk (0 meaning the agent never moved, 1 meaning the agent made it to the end), the number of times neighbors to a current node did not include the A* agent (the agent dying) and the number of times the A* agent fell below the level chunk's lowest point (falling off screen).

We collected these summarizing statistics across one-hundred sampled level chunks for each of the original three game graphs. 
This allowed us to derive a distribution over these values for the original games. 
However, given the time cost of simulating an A* agent,  we did not wish to simulate one-hundred level chunks for each game graph during the search process for every heuristic call.
Instead, we simulate only five level chunks and compared the median, first quartile, and third quartile of this five sample distribution and the existing one-hundred sample distributions in terms of the collected metrics.
In this case the goal for the system is to minimize the distance between this representation of value and the original games (as we assume the original games have appropriate challenge).
Thus if the median, first quartile, and third quartiles exactly match any of the three original existing games this returns a 1.0, with the linear distance taken otherwise.
This means that the value metric is the only metric that does not involve any other generated games previously output by the system, as there is no way to ensure the previously generated games have reasonable challenge.

Taken together then the maximum heuristic value is 3.0, with a maximum of 1.0 coming from the novelty, surprise, and value metrics.
An ideal game then would be nothing like any of the original input games or previously output games, while still being roughly as challenging.

\subsection{Game Graph to Unity Project}

We chose the game engine Unity for our study.
We wrote a simple Python script to translate game graphs into Unity projects so that they could be played. 
Each node associated with at least one animation fact became a GameObject, Unity's representation for all game entities. 
All of the rules were translated into one massive C\# script, made up of a long sequence of IF statements (usually several dozen per game).
We chose a keyboard button for the rules that required input.
For the level structure, the script converted all level design information of the game graph into a level graph. 
It then queried the level graph until it output a level that could be completed by our A* agent.
We used the same Unity camera behavior for all the generated games.
We made this choice for simplicity's sake, and to ensure human players could always see the player avatar, which was not a given with this system.

\section{Human Subject Study}

\begin{figure}[tb]
	\centering
	\includegraphics[width=\columnwidth]{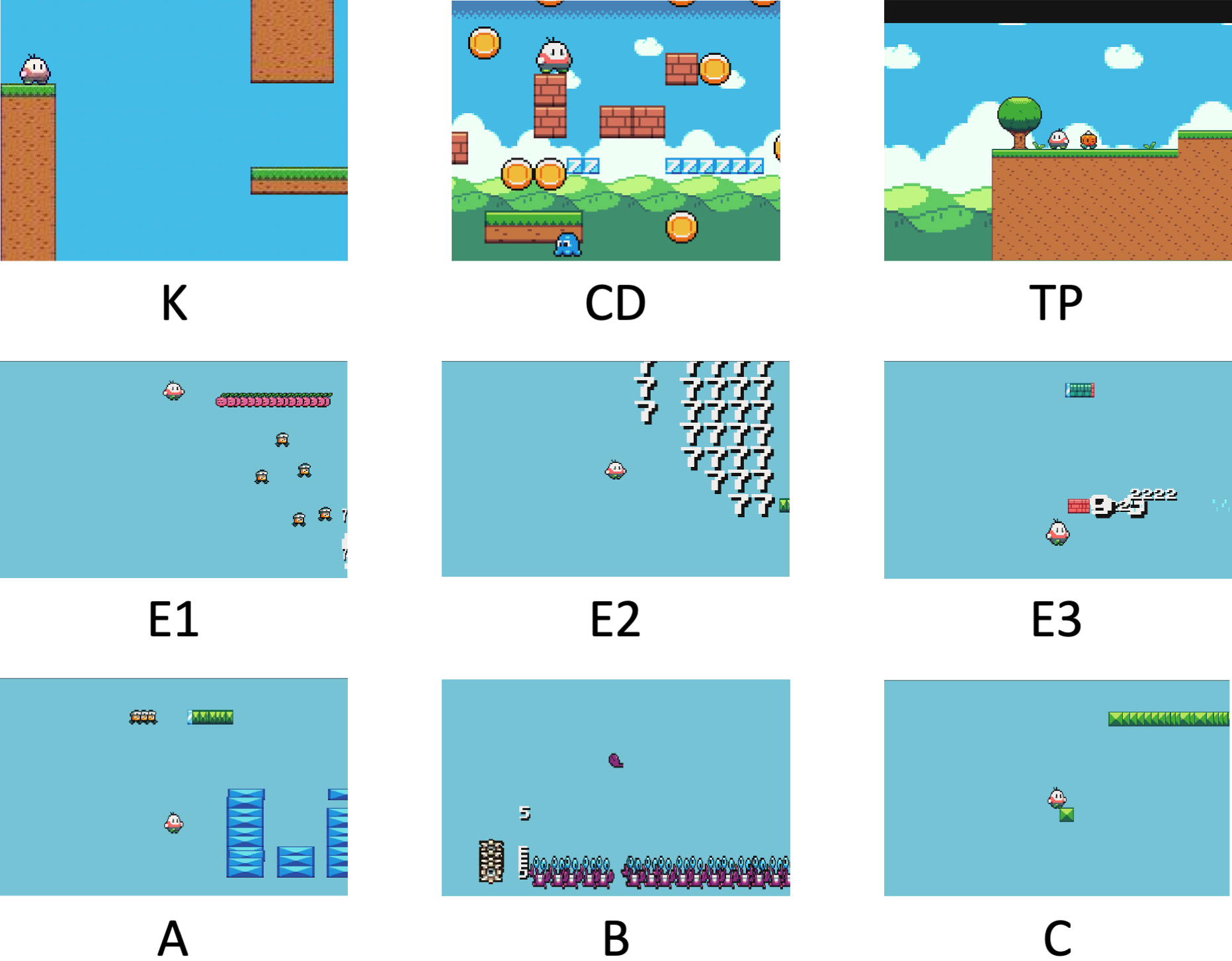}
	\caption{The first screen of the nine study games. The top row represents the human-produced games, the second row the CE games, and the final row the CC baseline games.}
	\label{fig:allGames}
\end{figure}

In this section, we discuss the human subject study that we designed to evaluate this spritesheet-based, conceptual expansion game generator.
We first discuss the nine games created for the study in terms of the three types of game generators: human game designers, our conceptual expansion approach, and the three existing combinational creativity approaches. 
We visualize the first screen of each game in Figure \ref{fig:allGames}, where each row represents one of these three types, in order.
The games are discussed in more detail below.
After the games, we discuss the process participants went through in the study.

\subsection{Human Games}

We contacted seven human game designers to make games for this study.
We asked these designers to create a game under the same constraints as the automated approaches, in terms of using the same sprites and producing a platformer game.
Three designers agreed to take part, each producing one game.
Due to our involvement in this process, a separate video game playing expert without knowledge of this study outside of these contracts verified the games were sufficient and met our required constraints.
We did not see the human-made games until the study began.
We highlight each designer and their game in the order in which the games were completed below.
Information about the game designers and games was collected after the designers submitted their games and had been approved by the third party.

\textbf{Kise (K): }The first game designer goes by Kise and described herself as a hobbyist game developer.
She developed her game in javascript, and took inspiration from the game Super Meat Boy.
The player retains acceleration in the x-dimension, leading to a ``slidey'' feeling.
The game is playable at http://guzdial.com/StudyGames/0/.
\textbf{Chris DeLeon (CD): }The second game designer was Chris DeLeon. 
He described himself as an ``online educator''.
His javascript game is much more of a puzzle game, with the goal being to collect all of the coins (some hidden in blocks) without falling off the screen or without a blue creature touching you.
One can find more from Chris DeLeon at ChrisDeLeon.com.
The game is playable at http://guzdial.com/StudyGames/1/.
\textbf{Trenton Pegeas (TP): }The third game designer was Trenton Pegeas.
When asked to describe himself he stated he'd ``like to classify as indie, but closer to hobbyist''.
He created his game in Unity.
He also created the most complex game, both in terms of creating the largest level and including the most sprites and entities in his game.
The player follows a fairly linear path winding along the level from their starting position, interacting with different entities until they make their way to a goal. 
One can find more from Trenton Pegeas at https://twitter.com/Xist3nce\_Dev. 
The game is playable at http://guzdial.com/StudyGames/2/.

\subsection{Conceptual Expansion Games}

We generated three conceptual expansion games (shortened to ``expansion games'') for this study.
After each conceptual expansion game graph was generated it was added to the knowledge base of existing games for the system, which impacted the novelty and surprise portions of the heuristic.
Thus the second game had the first game in its knowledge base and the third game had both the first and second games in its knowledge base. 
Below, we briefly describe all three games in the order they were generated.
The expansion games outperformed the combinational creativity baseline games described below in terms of the final heuristic values for the novelty metric, which is surprising given that they had additional points of comparison.
This is likely due to the larger output space of conceptual expansion.

\textbf{Expansion Game 1 (E1): }The first expansion game generated by the system, and therefore notably the only one based solely on the existing games, is not much of a platformer.
Instead the system produced something like a surreal driving or flying game, with the player moving constantly to the right and needing to avoid almost all in game entities since the player dies if it touches them.
The game is playable at http://guzdial.com/StudyGames/3/.
\textbf{Expansion Game 2 (E2): }The second expansion game, which included the prior game in its knowledge base for the novelty and surprise measures, is also not much of a platformer.
If the player moves left or right they slowly go in that direction while gently falling along the y-axis.
If the player goes up they shoot up quickly, meaning that the player has to balance going up to a set rhythm, somewhat like the game Flappy Bird.
If the player goes too high or too low they die and if they stay still they die.
There is nothing to dodge in this game, meaning that once the player gets the rhythm of hitting the up arrow or space bar it is a simple game. 
The game is playable at http://guzdial.com/StudyGames/4/.
\textbf{Expansion Game 3 (E3): }The third expansion game, which included the prior two games in its knowledge base for the novelty and surprise measures, was even more like Flappy Bird.
The player normally falls at a constant speed. 
The avatar moves slowly left or right when the player hits the left or right arrow.
If the player hits the up arrow or space bar the avatar shoots up, with the inverse happening if the player hits the down arrow.
The game is largely the same as the second game except for the more surreal level architecture and an area of ``anti-gravity'' (as dubbed by a study participant) whenever the player is above or below a patch of heart sprites.
The player stretches when they go up or down, but nothing animates otherwise.
The game is playable at http://guzdial.com/StudyGames/5/.

\subsection{Baseline Games}

To determine the extent to which conceptual expansion was better suited to this task compared to existing combinational creativity approaches, we included one game made by each of the three combinational creativity approaches highlighted in the related work (amalgamation, conceptual blending, and composition adaptation).

In all cases we employed a similar process to conceptual expansion search, only instead of sampling from the space of possible conceptual expansions we instead sampled from the particular output spaces of each approach. 
Essentially, we employed our heuristic as a means of choosing a single valid output from each approach in place of existing domain-free heuristics or constraints \cite{fauconnier2001conceptual,ontanon2010amalgams}.
The only change we made to these existing approaches was in allowing for more than two inputs, but given that all three approaches employed the same mapping as the conceptual expansion we largely side-stepped this issue.
We chose to include a single game from each approach as each approach has been historically designed to only produce one output for every input. 
We briefly highlight each game below.

\textbf{Amalgamation (A): }The amalgamation process meant that whole existing game graph nodes were used for each node of the proto-game graph.
Given the small size of valid outputs we were able to search the entirety of the output space to find the output (amalgam) that maximized our heuristic.
This led to perhaps the second most platformer-like of the generated games, with the player largely having the physics of kirby from Kirby's Adventure, though slightly broken due to some nodes from the Kirby game graph not being included.
Because of the Kirby-esque physics it is almost impossible to lose this game.
The game is playable at http://guzdial.com/StudyGames/6/.
\textbf{Conceptual Blending (B): }The conceptual blend combined as much of the mapped existing game graph nodes as possible for each node of the proto-game graph.
The valid output space for conceptual blending was roughly five times the size of the amalgamation space, but we were  able to search it completely and return the best output (blend) according to our heuristic.
This led to a game strikingly similar to the second and third expansion game, except with the notable difference that at the end of every frame the player transformed into a crab arm sprite.
Thanks to a serendipitous mapping that led to treating a crab sprite as ground this led to a surprisingly semantically coherent (if strange) game. The game is playable at http://guzdial.com/StudyGames/7/.
\textbf{Compositional Adaptation (C): }The compositional adaptation approach led to a game that we believe was the most like a standard platformer, given that it was essentially a smaller version of the amalgam game.
This is due to the fact that the output size is fixed for this problem, while the majority of the expressivity of compositional adaptation is in its ability to create output of varying sizes, thus the space of outputs was roughly the size of the amalgamation output space. 
The game again used the system's understanding of Kirby physics, but without the amalgamation requirement to incorporate as much information as possible it is more streamlined.
The game is essentially a series of small, very simple platforming challenges. 
The only strange thing was the random patches of numbers floating in the air, having had decorative elements from other games mapped onto them.
The game is playable at http://guzdial.com/StudyGames/8/.

\subsection{Human Subject Study Method}

The study was advertised through social media, in particular Twitter, Facebook, Discord, and Reddit. 
Participants were directed to a landing page where they were asked to review a consent form, and then press a button to retrieve an ID, the links to the three games they would play, and a link to the post-study survey. 
Each player played one human game, one expansion game, and one baseline game in a random order. 
Players were asked to give each game `` a few tries or until you feel you understand it''.
After this the player took part in the post-study survey.

The participants were asked to rank the games across the same set of experiential features employed in \cite{guzdial2016game}: fun, frustration, challenge, surprise, ``most liked'', and ``most creative''. 
Once the participant finished the ranking questions they were asked whether each game was made by a human or AI.
Participants had been informed ``some'' of the games were made by humans and ``some'' by AI, but not which was which.
This was made easier as all the AI-generated games were made in Unity, while two of the three human games were made in javascript. Following these questions participants were asked a series of demographic questions, concerning how often they played games, how often they played platformer games, how familiar they were with video game development, how familiar they were with artificial intelligence, their gender, and age. 

\subsection{Human Subject Study Results}

One-hundred and fifty-six participants took part in this online study.
Of these, one-hundred and seventeen identified as male, twenty-seven as female, six as non-binary, and the remaining six did not answer the gender question.
Eighty-eight of the participants placed themselves in the 23-33 age range, Forty-one in the 18-22 age range, and twenty-five in the 34-55 age range.
There was a strong bias towards participants familiar with video game design and development, with one-hundred and eight participants ranking their familiarity as three or more on a five point scale.
Similarly, one-hundred and twelve of the participants played games at least weekly, though only eighty-two played platformer games at least monthly.
These results suggest that this population should have been well-suited to evaluating video games.
There was also a strong bias towards the participants being AI experts with eighty-one selecting four or more out of five, with an additional thirty-two selecting three. 
Thus these participants should have been well-suited to evaluating AI-generated video games, and skilled at differentiating between human and AI generated games.

Our principle hypothesis was that the conceptual expansion games would outperform the baseline games in terms of value and surprise, with the human games considered a gold standard.
We do not include novelty as we have an objective, quantitative measure for novelty, while both the value and surprise components of the heuristic depended on approximations of human perception.
We identify the following hypotheses for the purpose of interpreting our results.
\begin{enumerate}
    \item[H1.] The expansion games will be more similar to the human games in terms of challenge in comparison to the baseline games.
    \item[H2.] The expansion games will be overall the most surprising.
    \item[H3.] The expansion games will be more like the human games in terms of the other attributes of game value.
    \item[H4.] The expansion games will be more creative than the baseline games. 
\end{enumerate}

\noindent
Hypothesis H1 essentially mirrors the measure we designed for value, creating games that are challenging in the ways that the existing, human-designed games are challenging.
H2 in turn mirrors the measure we designed for surprise, which aims to be as or more surprising than the human-designed games.
H3 is meant to determine the importance of the choice of heuristic when it comes to conceptual expansion, whether the approach carries over aspects of the input data not represented in the heuristic (e.g. all of the input games are fun, but the heuristic doesn't attempt to measure that).
H4 is meant to determine if conceptual expansion on its own reflects the human creative process better than the existing combinational creativity approaches.
In the following sections we break down different aspects of the results in terms of these hypotheses.

Participants had no issue differentiating between the human and artificially generated games outside of a few outliers.
We therefore do not include these results in our analysis.

\subsection{Quantitative Results By Author Type}

The first step was to determine the aggregate results of participant's experience with the games, broken into the three types we identified above (human, expansion, baseline).
One can consider this as separating the games into different groups according to the nature of their source or author as long as one treats the baseline combinational creativity approaches as part of the same class of approach as in \cite{guzdial2018combinatorial}.
By bucketing the games into one of three author types there were essentially six possible conditions participants could be placed in, based upon the order in which they interacted with games from these three types. 
Because of randomly assigning participants, the one-hundred and fifty-six participants were not evenly spread across these six conditions. 
The smallest number was in the category where participants played a human game first, an expansion game second, and a baseline game third, with only nineteen participants.
We randomly selected nineteen participants from each of the six categories, and use these one-hundred and fourteen results for analysis in this section.
We sample only once, as multiple rounds of sampling would greatly increase the probability of false positives \cite{sullivan2016common}.

\begin{table}[tb]
\begin{center}
\caption{A table comparing the median experiential ranking across the different games according to author type.}
\begin{tabular}{|l|c|c|c|} 
 \hline
  & Human & Expansion & Baseline\\
  \hline
Fun & 1st & 3rd & 2nd\\
  \hline
Frustrating & 2nd & 2nd & 2nd\\
   \hline
Challenging & 1st & 2nd & 3rd\\
   \hline
Surprising & 3rd & 1st & 2nd\\
  \hline
Liked & 1st & 3rd & 2nd\\
  \hline
Creative & 2nd & 2nd & 2nd\\
  \hline
\end{tabular}
\end{center}
\end{table}

We summarize the results according to the median value of each ranking in Table 1.
We report median values as the rankings are ordinal but not necessary numeric.
We discuss the results more fully below and give the results of a number of statistical tests.
Due to the number of hypotheses and associated tests we use a significance threshold of $p<0.01$, this is roughly equivalent to the Bonferroni correction of $p=0.02$ for each feature.
For all comparisons we ran the paired Wilcoxon Mann Whitney U test, given that these are non-normal distributions.

Hypothesis H1 posits that the expansion games should be ranked more like the human challenge rankings than the baseline challenge rankings. 
The median values in Table 1 give some support to this.
Further, there are significant differences between the human and baseline challenge rankings ($p=9.62e{-6}$), and between the expansion and baseline challenge rankings ($p=7.61e{-9}$).
However, we are unable to reject that the human and expansion challenge rankings come from the same underlying distribution ($p=0.041$).
H1 is supported by these results.

H2 suggests that the expansion games should be more surprising than either of the other two types of games. 
The median ranking was first (most surprising) for the expansion games, lending credence to this.
Statistically, we found that the expansion games were ranked significantly higher in terms of surprise than both the human ($p=1.95e{-13}$) and baseline ($p=4.83e{-6}$) rankings.
Further, the baseline games were ranked as more surprising than the human game rankings ($p=3.81e{-9}$).
H2 is thus supported.
These results suggest that combinational creativity approaches may generally produce more surprising results than human generated games for this particular problem. 
However, the conceptual expansion approach led to the most surprising output overall.
Part of the reason for these results may be that the expansion games were the least platformer-like. 
However, the participants were not told to expect platformer games. 
Further, the expansion and baseline games shared the same input and heuristic, meaning they were not biased to produce platformer-like games. 

H3 suggested that the expansion games would be ranked more like the human games for other elements of game value, which we investigate in terms of the fun, frustrating, and most liked results.
In terms of fun the median rankings do not support this hypothesis, with expansion ranked third more frequently than the baseline ranking.
The statistical test was unable to reject the null hypothesis that the baseline and expansion fun rankings came from the same distribution ($p=0.39$).
But the human rankings were significantly higher than both the baseline ($p=1.03e{-8}$) and expansion rankings ($p=1.23{e-9}$).
For the frustrating question all of median rankings were exactly the same.
However, the statistical tests allowed us to compare the distributions, with the expansion games ranked as significantly more frustrating overall in comparison to both the human ($p=1.61e{-3}$) and baseline ($p=1.84e{-4}$) rankings.
The human and baseline frustrating rankings were too similar to reject the that they arose from the same distribution ($p=0.25$).
Finally, the most liked results parallel the fun results, with the human games ranked significantly higher than baseline ($p=1.24e{-9}$) and expansion ($p=1.32e{-9}$) rankings, but with the baseline and expansion rankings unable to reject the null hypothesis that they arose from the same distribution ($p=0.66$).
This evidence does not support H3, with the expansion games found to be roughly as fun and liked as the baseline games, and more frustrating than all other games.

H4 states that the expansion games will be viewed as more creative than the baseline games.
However, as with frustration, the median ranking of all of these games suggests that the participants evaluated the games equivalently on this measure.
Unlike frustration, the statistical tests are unable to pick apart these results, with no comparison able to reject that they arose from the same distribution ($p>0.7$).
Thus H4 is not supported by these results.

\subsection{Quantitative Results By Game}

In this section, we examine the survey results by each game individually. 
We do this for a number of reasons.
First, the human and baseline games have three very different ``authors'', with the human games created by three humans and the baseline games created by three combinational creativity approaches.
Second, the conceptual expansion games build off one another in terms of surprise and novelty, which cannot be captured in aggregate. 
Third, humans were ranking individual games, not the sources of these games. 
Finally, this allows for a more nuanced view of our results.

For this section we instead analyze the results in terms of each game individually. 
Taken this way, there are 27 distinct categories in terms of a random selection of three of these nine games, and this is without taking into account the ordering of these games.
Given the random assignment the results are too skewed to run statistical tests for these categories (there are only two participants who interacted with Kise's game (K), Expansion game 2 (E2), and the adaptation game (A), as an example).
Instead we treat these rankings as ratings, given that a multi-way ANOVA cannot find any significant impact from ordering on any of the experiential factors ($p>0.01$).
In this case we determine the game with the fewest number of results (the blend with only forty-three participants), and randomly select this number of ratings for each of the nine games, for a total of three-hundred and eighty-seven ratings per experiential measure (compare this to the four-hundred and sixty eight total ratings available).
We employ these ratings for the analysis in this section, due to the number of comparisons we employ the same conservative significant threshold of $p<0.01$, and we use the unpaired Wilxocon Mann Whitney U test.
We present the results in Table 2 in terms of median rating per game.

\begin{table}[tb]
\begin{center}
\caption{A table comparing the median ratings across each measure for each game. We mark median ratings of ``1'' in bold to make the table easier to parse. ``1'' means best/most, as in most fun, most liked, etc. and ``3'' means worst/least.
}
\begin{tabular}{|l|c|c|c|c|c|c|c|c|c|} 
 \hline
  & K & CD & TP & E1 & E2 & E3 & A & B & C\\
  \hline
Fun & \textbf{1} & \textbf{1} & \textbf{1} & 2 & 3 & 3 & 2 & 2 & 2\\
  \hline
Frustrating & 2 & 2 & 3 & 2 & \textbf{1} & \textbf{1} & 2 & 2 & 2\\
  \hline
Challenging & \textbf{1} & 2 & 2 & 2 & 2 & 2 & 3 & 3 & 2\\
  \hline
Surprising & 3 & 3 & 3 & 2 & \textbf{1} & \textbf{1} & 2 & 2 & 2\\
  \hline
Liked & \textbf{1} & \textbf{1} & \textbf{1} & 2 & 3 & 2 & 2 & 3 & 2\\
  \hline
Creative & 2 & 2 & 2 & 2 & 2 & 2 & 2 & 2 & 2\\
  \hline
\end{tabular}
\end{center}
\end{table}

The ratings give more nuanced results in terms of hypothesis H1. 
Hypothesis 1 states that the expansion games (E1, E2, and E3) should be rated more like the human games (K, CD, and TP) than the baseline games (A, B, and C) in terms of challenge.
For E2 and E3, there was no significant difference between their challenge ratings and the challenge ratings of any of the human games, but there were significant differences found between them and the challenge ratings of the amalgamation (A) and blend (B) games.
However, E2 and E3's challenge ratings did not differ significantly from the challenge ratings of the composition game (C). 
E1's challenge ratings only differed significantly with Kise's game (K), they were too like the other challenge ratings to differ significantly for any of the other games. 
Thus, H1 is supported.

Hypothesis H2 suggests that the expansion games (E1, E2, and E3) should be rated more surprising than either the human games or baseline games.
E2 and E3 both had significantly higher surprise ratings than all but the amalgamation (A) and blend (B) games.
However, while not significant according to the unpaired Wilxocon Mann Whitney U test, their median surprise ratings were still higher than these two games, as demonstrated in Table 2.
Comparatively, E1 only had significantly higher surprise ratings than the three human games.
There was no significant difference between E1's surprise ratings and the surprise ratings of any of the baseline games. 
This follows from the way the heuristic modeled surprise, which would have pushed E2 and E3 to be surprising than the expansion game(s) output before them. 
Thus, H2 is supported.

Taken in aggregate the results in the prior section did not support H3, which stated that the expansion games would be evaluated more like the human games than the baseline games for measures of game value outside of challenge.
Looking at the results by game the picture becomes more nuanced.

In terms of the \textit{fun} ratings,the games follow their aggregate rankings, but E1 stands out as an outlier from the other two expansion games. 
In fact, E1 is found to be rated significantly more fun compared to E2 ($p=4.22e{-8}$), E3 ($p=3.601e{-5}$), and the blend game ($p=2.49e$).
However, we cannot reject that this rating distribution comes from the same rating distribution as the amalgam ($p=0.03$) and composition ($p=0.27$) games.
Similarly, there is no significant difference between E1 and CD ($p=0.16$), while both the remaining human games are rated significantly more fun ($p<0.001$).
This suggests that E1 is more fun than the blend game, and roughly as fun as Chris DeLeon's game (CD), the amalgam game (A), and the composition game (C). 

For the \textit{frustrating} ratings, E2 and E3 stand out as outliers from the remaining games, being rated as the most frustrating the majority of the time.
These games are clearly the cause of the expansion games being found to be significantly more frustrating when taken in aggregate compared to the other types of games, which individual tests confirm ($p<0.01$).
However, there is no significant difference comparing the E1 frustrating ratings and any of the other games.
This suggests E1 is about as frustrating as all of the non-expansion games. 

For the \textit{liked} rating, both E1 and E3's ratings differ from the aggregate expansion games median rating of 3.
While E2 is rated significantly lower than all other games on this measure ($p<0.01$), both E1 and E3 are not.
E1 is rated significantly less well-liked in terms of this measure than Trenton Pegeas' Game (TP), and significantly higher rated than the blend game, with no significant difference from any of the other non-expansion \textit{liked} ratings.
Comparatively, E3 is rated significantly lower than all three human games, with no other significant differences comparing any of the other \textit{liked} rating distributions.

Overall, these results suggest weak support for H3, given that at least E1 seemed to be more like at least one of the human games (CD) in terms of fun, frustration, and likeability.
This points to a potential issue with the heuristic, which may have biased games towards surprise and novelty at the cost of other positive elements of the input games that impacted these experiential features.
The effect of this would have been compounded for E2 and compounded again for E3 given the way the previously output games impacted the novelty and surprise measures.
We discuss this further in the next section.

Given H4, we would expect to find the expansion games rated more creative compared to the other games. 
However, taken individually, all the games had equal median values, pointing to the same issue identified in \cite{guzdial2016game} that untrained participants appear to have difficulty consistently evaluating creativity.
However, both E3 and the amalgam are rated significantly more creative than Kise's game ($p<0.01$). 
However, they do not differ significantly from any of the other games ratings.
Thus we can only consider there to be weak support for H4.

\subsection{Discussion}

The results suggest support for H1 and H2, specifically that players view the games output by the conceptual expansion game generator as similar to the human-made games in terms of challenge and more surprising than human-made games or other combinational creativity games.
This seems to follow from our choice of heuristic, which involved very specific measures for value, surprise, and novelty. 
Notably, all the combinational creativity approaches used the same heuristic to search their spaces of potential output, but the output space of conceptual expansions had higher measures of these values.
This had both positive and negative consequences. 
While the expansion games outperformed the baselines in terms of challenge and all games in terms of surprise, the latter two games seemed to gain surprise at the expense of other measures of game value besides challenge (specifically being less fun, more frustrating, and less well-liked). 
This could in part be due to the nature of the heuristic, where novelty and surprise made up the majority of the output value, or due to the heuristic only focusing on challenge as a representation of value.
One of the benefits of combinational creativity is that it retains good features from the inputs to the combination \cite{fauconnier1998conceptual}. 
However, our heuristic appears to have been unbalanced, especially in the later outputs, which had compounding effects in terms of novelty and surprise from the earlier output. 
For applications of conceptual expansion in domains where retaining more of the input features is desirable, we intend to avoid this compounding effect in future work. 

We believe our optimization approach likely impacted our output games. 
We employed a greedy stochastic optimization, and thus ``E1'' is likely a local optima according to our heuristic.
While referencing prior output games in the heuristic allowed our approach to avoid the same local optima, all the expansion games are still similar.
In the future we hope to explore more optimization approaches to encourage diversity, such as Quality Diversity~\cite{gravina2019procedural}, a group of evolutionary optimization methods that attempt to increase a population's fitness while maintaining variety within the members of the population. 
These methods could allow for the production of multiple games that demonstrate more distinct mechanics. 

While improvements to the optimization approach and heuristic could benefit future applications of conceptual expansion to novel game generation, these are both open research problems. 
In particular, we believe that a general heuristic for game quality will remain an open challenge in PCG research \cite{togeliusSurvey}.
To sidestep these problems we intend to focus on a future system in which a user can guide the conceptual expansion search process \cite{liapis2016mixed}.
A mixed-initiative system could allow users to pick and choose games from the space of conceptual expansions, allowing for a flexible and subjective notion of game quality and driving the exploration of the space.

\section{Conclusions}

In this paper, we presented an approach employing conceptual expansion of game graphs for automated game generation.
We described our game graph representation, its component parts, and how we automatically learned game graphs for three existing games. 
We then discussed our approach for deriving novel games from these learned game graphs by representing the novel games as combinations of existing games, and searching the space of possible combinations.
Our output games outperformed existing combinational approaches in terms of the heuristic we employed.
We ran a human subject study comparing our output games with games output by existing combinational approaches, and games authored by humans. 
Participants ranked one of our of our output games as similar to one human-designed game in terms of fun, frustration, and likeability.
Our results indicate conceptual expansion can lead to valuable output when applied to learned representations of games, but further work is required to improve the approach's consistency.

\section*{Acknowledgements}

This material is based upon work supported by the National Science Foundation under Grant No. 1002748 and Grant No. 1525967. Any opinions, findings, and conclusions or recommendations expressed in this material are those of the authors and do not necessarily reflect the views of the National Science Foundation.

\bibliographystyle{IEEEtran}
\bibliography{main}

\end{document}